\documentclass{SCIS2025}

\usepackage{graphicx} 
\usepackage{hyperref}
\usepackage{multirow}
\usepackage{array}
\usepackage{subcaption}
\usepackage{xcolor}
\newcommand{\blue}[1]{{\color{black}#1}}

\usepackage{enumitem}

\newcommand{\zl}[1]{{\color{black}#1}}

\newcommand{\ignore}[1]{}

\usepackage{xspace}
\newcommand{\oursys}{\texttt{SAGE-Health}\xspace}
\usepackage[numbers,sort&compress]{natbib}
\usepackage{etoolbox}
\begin{document}


\title{Generative AI for Healthcare:\\ Fundamentals, Challenges, and Perspectives}



\author[1]{Gang Chen}{}
\author[3]{Changshuo Liu}{}
\author[4] {Gene Anne Ooi}{}
\author[5]{Marcus Tan}{}
\author[2]{Zhongle Xie}{}
\author[1,2]{\\ Jianwei Yin}{}
\author[5]{James Wei Luen Yip}{}
\author[2]{Wenqiao Zhang}{}
\author[3]{Jiaqi Zhu}{jiaqi77@nus.edu.sg}
\author[1,2]{Beng Chin Ooi}{}

\address[1]{College of Computer Science and Technology, Zhejiang University, Hangzhou 310027, China}
\address[2]{College of Software Technology, Zhejiang University, Ningbo 315100, China}
\address[3]{School of Computing, National University of Singapore, Singapore 117417}
\address[4] {Singapore General Hospital, Singapore 169608}
\address[5]{National University Hospital, Singapore 119074}

\abstract{
\begin{abstract}

Generative Artificial Intelligence (GenAI) is taking the world by storm.
It promises transformative opportunities for advancing and disrupting existing practices, including
healthcare.
From large language models (LLMs)  for clinical note synthesis and conversational
assistance to multimodal systems that integrate medical imaging, electronic health records (EHRs),
and genomic data for decision support, GenAI is transforming the practice of medicine and the delivery of healthcare, such as diagnosis and personalized treatments, with great potential in reducing the cognitive burden on clinicians, thereby improving overall healthcare delivery.
However, GenAI deployment in healthcare requires an in-depth understanding of healthcare tasks and what can and cannot be achieved.

In this paper, we propose a data-centric paradigm in the design and deployment of GenAI
systems for healthcare.
Specifically, we
reposition the data life cycle by making the medical data ecosystem as the foundational substrate for generative healthcare systems.
This ecosystem is designed to sustainably support the integration, representation, and retrieval of diverse medical data and knowledge.
With effective and efficient data processing pipelines, such as semantic vector search and contextual querying, it enables GenAI-powered operations for upstream model components and downstream clinical applications. Ultimately, it not only supplies foundation models with high-quality, multimodal data for large-scale pretraining and domain-specific fine-tuning, but also serves as a knowledge retrieval backend to support task-specific inference via the agentic layer.
The ecosystem enables the deployment of GenAI for high-quality and effective healthcare delivery.
\end{abstract}
}


\maketitle

\section{Introduction}
\label{sec:intro}

The rapid advancement of Generative Artificial Intelligence (GenAI) is creating unprecedented opportunities for healthcare.
Large language models (LLMs)~\cite{wu2024pmc,singhal2023large,singhal2025toward,labrak2024biomistral} are being applied to tasks such as clinical note synthesis and conversational assistance, while multimodal intelligence systems~\cite{wang2022medclip,wang2022multi,bannur2023learning,yang2024unlocking,lai2024carzero,lin2025healthgpt,li2025eyecaregpt} that integrate medical imaging, electronic health records (EHRs), and genomic data are emerging as powerful tools for sophisticated decision support.
Such developments highlight the potential of GenAI to accelerate diagnosis, enable personalized treatments, and alleviate the cognitive burden on clinicians.

\begin{figure}[t]
\centering
  \begin{minipage}[b]{0.48\textwidth}
    \centering
    \includegraphics[width=0.7\linewidth]{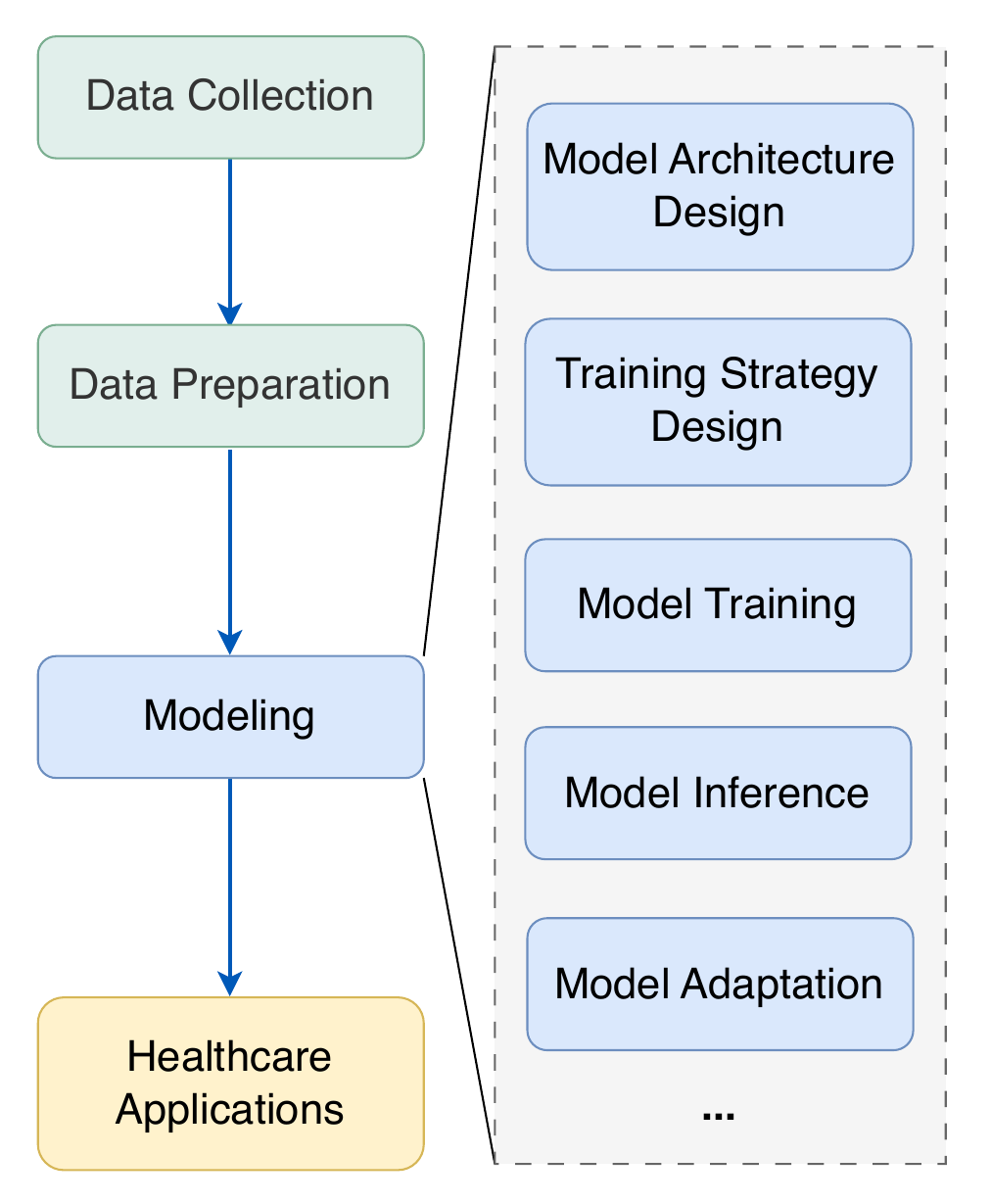}
    \par\vspace{2pt}\footnotesize (a)Static model-centric pipeline
  \end{minipage}\hfill
  \begin{minipage}[b]{0.5\textwidth}
    \centering
\includegraphics[width=0.93\linewidth]{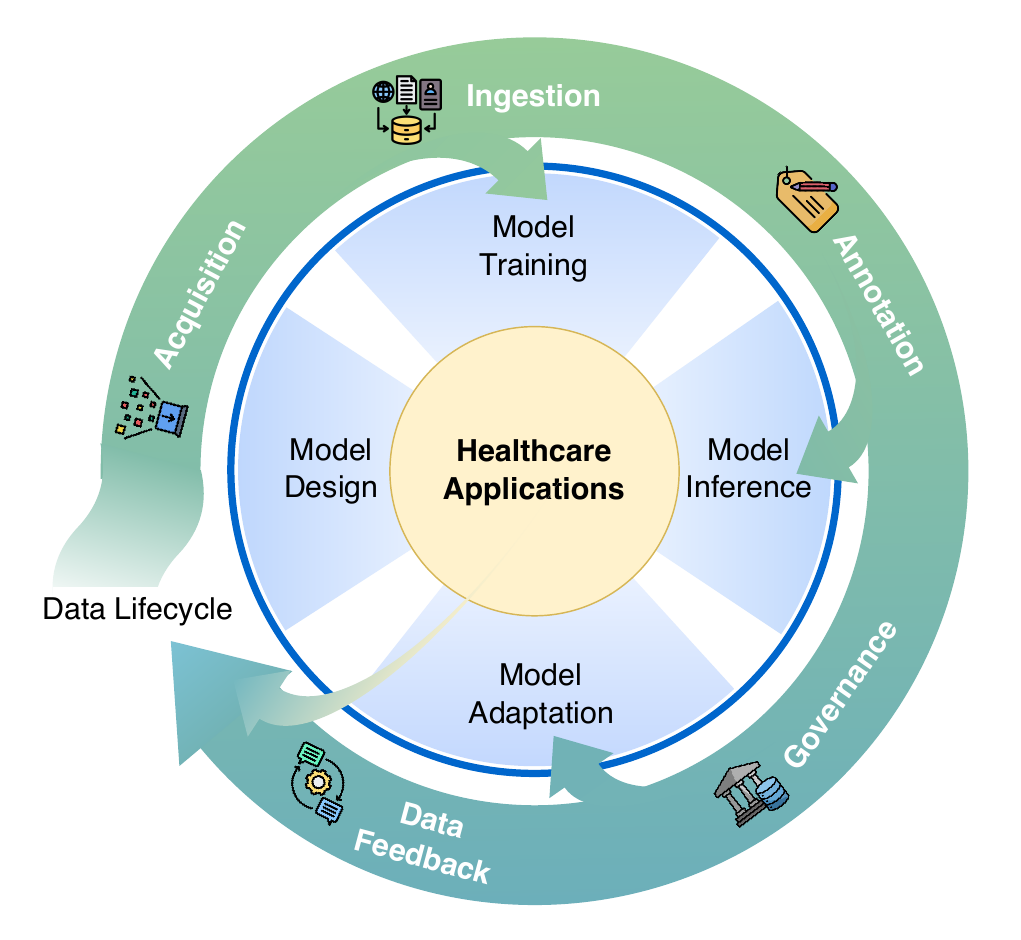}
    \par\vspace{2pt}\footnotesize (b)Evolving data-centric ecosystem for healthcare applications
  \end{minipage}
\caption{Evolution to co-evolving healthcare GenAI with a sustainable data ecosystem. }
\label{fig:background}
\end{figure}

While model innovation in GenAI keeps ``tachytely'' evolving, driven by advances in large-scale pretraining, in-context learning, and architectural scaling~\cite{bommasani2021opportunities,achiam2023gpt}, sustained progress depends on a fundamental systemic transformation that treats data as coequal with models.
As illustrated in Figure~\ref{fig:background},
the transformation entails a shift from static, model-centric pipelines to evolving, data-centric ecosystems. 
Within such ecosystems,
the performance of foundation models depends 
heavily on the quality, quantity, and diversity of their data~\cite{ooi2024neurdb,fries2022bigbio,Singa2015,wang2016database}, and relies on effective large-scale pretraining, domain-specific fine-tuning, and retrieval augmented generation.
The trajectory of the GPT series exemplifies this dependency, in which progressively larger and higher quality datasets, 
and effective prompt engineering, have
resulted in better performance with
relatively similar model architectures.
%

Despite the central role of data in enabling GenAI, the current medical data landscape remains highly fragmented, heterogeneous, and siloed across institutions, modalities, and systems.
Clinical data, including structured EHRs, free-text notes, medical images, physiological signals, and genomic information, are often stored in disparate formats with inconsistent standards, limited interoperability, and restricted accessibility~\cite{Ooi2017, johnson2023mimic}.
This not only hampers model training and evaluation but also limits the scalability and generalizability of GenAI applications across real-world settings.
Furthermore, annotation bottlenecks, privacy regulations, and institutional policies severely constrain the availability of large-scale and high-quality datasets~\cite{price2019privacy}.
Recent efforts~\cite{huang2019clinicalbert,gu2021domain,yang2022gatortron,yang2022large} have introduced medical foundation models pretrained on domain-specific corpora (e.g., PubMed\footnote{\url{https://pubmed.ncbi.nlm.nih.gov/}}, MIMIC\footnote{\url{https://mimic.mit.edu/}}, or multi-hospital imaging archives),  as summarized in Figure ~\ref{fig:foundation}.
Unfortunately, many of existing models still struggle with incomplete supervision, data bias, and poor alignment with downstream tasks.
Other emerging solutions, such as federated learning~\cite{pfitzner2021federated, falcon}, synthetic data generation~\cite{giuffre2023harnessing}, and self-supervised representation learning~\cite{azizi2021big,ye2024continual} may be able to resolve some of the issues.
Nevertheless,
a robust, intelligent, and scalable data ecosystem for GenAI in healthcare remains elusive.

As opposed to
viewing data as a passive, one-off input to model-centric GenAI healthcare pipelines, we advocate a sustainable data ecosystem that co-evolves with the foundation models and clinical applications to enable reliable, scalable, and context-aware generative capabilities.
To realize this paradigm, we identify three interlocking challenges.
Firstly, the inherent fragmentation and heterogeneity of medical data present fundamental obstacles to effective utilization.
Secondly, comprehensive data management and governance frameworks remain largely aspirational rather than operational.
Lastly, current healthcare systems lack the essential infrastructure for data–model co-evolution, limiting the continual refinement and contextual adaptability.
Overall, the translation of GenAI from proof-of-concept demonstrations to practical clinical deployment depends fundamentally on realizing \textbf{Data Value} through three pillars: cohesive data integration across modalities, rigorous life cycle governance frameworks, and sophisticated infrastructure supporting data–model co-evolution.
Only through addressing these interconnected challenges can we bridge the gap between GenAI's theoretical potential and its transformative application in clinical practice.

To explore this vision, we propose a data-centric GenAI system for healthcare that fundamentally reconceptualizes the medical data life cycle. We position \oursys, a \textbf{\underline{S}}ustainable, \textbf{\underline{A}}daptive, and \textbf{\underline{G}}enerative \textbf{\underline{E}}cosystem for Healthcare, which serves as a comprehensive medical data ecosystem as the foundational substrate for generative healthcare systems, designed to sustainably integrate, represent, and retrieve diverse medical data and knowledge sources.
\oursys enables GenAI-powered operations through intelligent data management services, including semantic vector search and contextual querying capabilities. The architecture serves a dual purpose: supplying foundation models with high-quality, multimodal data for large-scale pretraining and domain-specific fine-tuning, while simultaneously functioning as a knowledge retrieval backend that supports task-specific inference via the agentic layer.
Moreover, the system incorporates a continuous improvement mechanism whereby user interactions and application-level feedback are systematically channeled back into the data ecosystem. This feedback loop enables dynamic data refinement and ongoing system evolution, ensuring that data quality, model performance, and clinical utility continuously enhance one another over time.
In summary, this integration establishes a self-reinforcing cycle that drives increasingly sophisticated and clinically relevant GenAI solutions for healthcare.

The structure of this paper is organized as follows.
Section~\ref{sec:Foundations} presents relevant background on the foundations of GenAI in healthcare. Section~\ref{sec:applications} reviews existing GenAI-powered applications across key healthcare scenarios.
In Section~\ref{sec:discussion}, we identify critical challenges and delineate the key features central to the generative healthcare systems. Section~\ref{sec:perspectives} outlines a conceptual roadmap for building GenAI-ready data ecosystems. 
We conclude the paper in Section~\ref{sec:conclusion} with a summary of insights and future directions.

\section{Foundations of Generative AI in Healthcare}
\label{sec:Foundations}

\begin{figure}[t]
    \centering 
\includegraphics[width=1\linewidth]{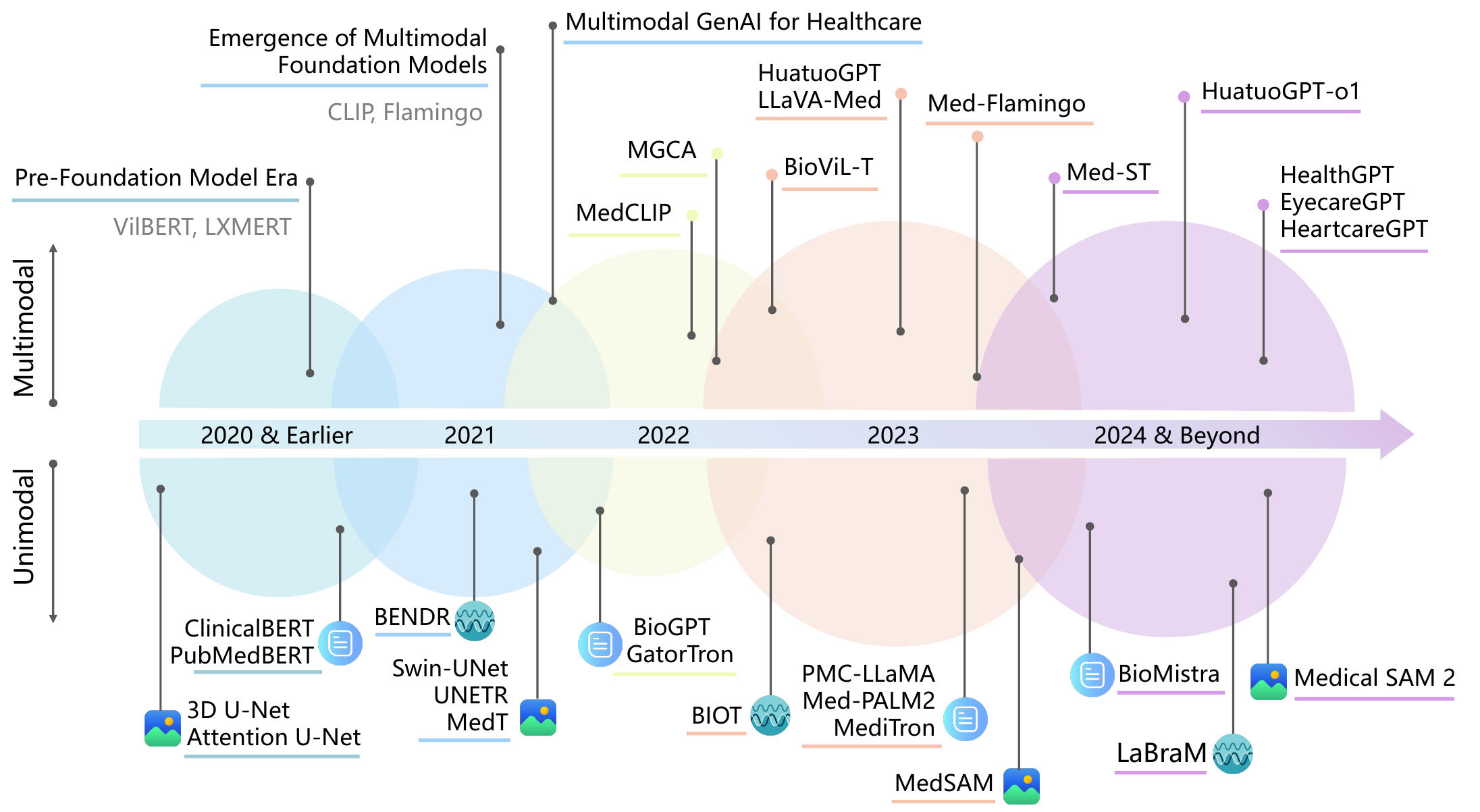} 
    \caption{Evolution of GenAI for healthcare: From the pre-foundation model era to domain-specific medical foundation models. }
    \label{fig:foundation}
\end{figure}

Many deep learning models have been proposed and 
\zl{deployed} for healthcare, including healthcare model zoos as the one in Apache SINGA~\cite{Singa2015}. 
They have been \zl{tailored} for specific regression \zl{or} prediction tasks and are often based on specific datasets.
Healthcare GenAI, making use of much larger amounts of heterogeneous data, provides \zl{multimodal understanding and generation across a wide range of clinical and biomedical scenarios}.
In this section, we shall briefly present the core building blocks of GenAI for healthcare, outlining the representative foundational models and infrastructures that underpin its development.
Figure~\ref{fig:foundation} provides a partial list of healthcare foundation models, unimodal and multimodal, proposed in recent years.

\subsection{Text‑centric Foundation Models}
With the advent of Transformer‑based architectures, 
Bidirectional Encoder Representations from Transformers (BERT)~\cite{devlin2019bert} has driven the rapid evolution of domain‑specific language models for clinical text understanding.
Early efforts, such as ClinicalBERT~\cite{huang2019clinicalbert} and BEHRT~\cite{li2020behrt}, extend BERT‑style pretraining to EHRs and clinical notes, significantly enhancing capabilities in information extraction, risk prediction, and event detection.
PubMedBERT~\cite{gu2021domain} further enhances \zl{domain} semantic representation by pretraining directly on biomedical literature from scratch, while GatorTron~\cite{yang2022gatortron,yang2022large} scales up both model size and training corpus, pushing clinical NLP toward \zl{higher} precision and \zl{broader} coverage.
For structured and temporal medical data, Med-BERT~\cite{rasmy2021med} targets longitudinal EHR sequences to enable dynamic health monitoring and risk stratification.
These models leverage transformer~\cite{vaswani2017attention} architectures to model complex temporal dependencies, thereby improving personalized medicine and clinical decision support.

Beyond BERT‑style encoders, a new generation of large‑scale, generative medical foundation models has emerged, shifting from bidirectional masked‑language modeling to instruction‑tuned, autoregressive architectures.
PMC-LLaMA~\cite{wu2024pmc} adapts the LLaMA framework through continual pretraining on biomedical literature and textbooks, achieving strong performance on medical question answering and reasoning.
Med‑PaLM~\cite{singhal2023large} and its successor Med‑PaLM 2~\cite{singhal2025toward} extend Google’s PaLM~\cite{chowdhery2023palm,anil2023palm} architecture with large‑scale instruction tuning, retrieval‑augmented reasoning, and ensemble refinement, \zl{achieving} state‑of‑the‑art performance on MedQA~\cite{jin2021disease} and \zl{the} United States Medical Licensing Examination (USMLE).


\subsection{Physiological Signal Foundation Models}

Physiological signal modeling has emerged as a critical component of generative AI in healthcare, driven by the need to extract clinically meaningful patterns from complex sequential data such as electrocardiograms (ECG) and electroencephalograms (EEG).
\zl{
Recent advances have been driven by transformer-based architectures and self-supervised pretraining, which enable scalable, task-agnostic feature learning from large volumes of biosignals.
}
For example, BENDR~\cite{kostas2021bendr} combined masked autoencoding and contrastive learning to enhance EEG representations across tasks and paradigms, while EEG2VEC~\cite{zhu2023eeg2vec} jointly optimized contrastive and reconstruction losses for robust local–global feature extraction, with the pretrained model serving as a feature extractor for downstream tasks.
Extending these ideas, ECGBERT~\cite{choi2023ecgbert} applied LLM-style masked modeling to ECGs, uncovering latent language-like structures for diverse diagnostic tasks with minimal fine-tuning.
EEGPT~\cite{wang2024eegpt} further advanced this paradigm as a pretrained transformer model designed for universal EEG feature extraction, incorporating spatio-temporal representation alignment and hierarchical processing to yield high-quality features for broad biomedical applications.

Moving toward greater generalizability, the Biosignal Transformer (BIOT)~\cite{yang2023biot} introduced a flexible encoder for cross-dataset learning, tokenizing heterogeneous biosignals (e.g., EEG, ECG, sensory data) into fixed-length segments to improve robustness across variable formats.
Similarly, the Large Brain Model (LaBraM)~\cite{jiang2024large} leveraged vector-quantized neural codes and masked code prediction for cross-dataset EEG pretraining, achieving state-of-the-art performance across diverse physiological signal modeling tasks.

\subsection{Medical Imaging Foundation Models}

In the medical imaging domain, architectural innovations have driven the shift from task‑specific solutions to generalizable foundation models, reshaping how visual representations are learned and transferred.
U‑Net~\cite{ronneberger2015u} has played a foundational role in medical image analysis, serving both as a benchmark architecture and a building block for many foundation models.
Originally introduced for biomedical segmentation, its encoder–decoder structure with skip connections enables precise localization from limited annotated data, making it highly adaptable across imaging modalities.
Building on this foundation, 3D U‑Net~\cite{cciccek20163d} extends the framework to volumetric CT and MRI.
Attention U-Net~\cite{oktay2018attention} incorporates attention mechanisms to selectively emphasize salient regions.
nnU‑Net~\cite{isensee2021nnu} automates architecture and hyperparameter optimization, establishing a task‑agnostic baseline for medical image segmentation.

With the rise of Vision Transformers (ViTs)~\cite{dosovitskiy2021image}, researchers have begun exploring their advantages over purely convolutional architectures, particularly their capacity for modeling long‑range dependencies and flexible contextual reasoning.
Medical Transformer (MedT)~\cite{valanarasu2021medical} proposes gated position-sensitive axial attention to improve training efficiency and contextual representation for medical images.
Additionally, hybrid models such as UNETR~\cite{hatamizadeh2022unetr} and Swin‑UNet~\cite{cao2022swin} integrate transformer encoders with U‑Net‑style decoders to combine global contextual reasoning with high‑resolution reconstruction.
More recently, Segment Anything Model (SAM)~\cite{kirillov2023segment,ravi2025sam} has emerged as a foundation model for universal image segmentation, and its medical adaptations, such as MedSAM~\cite{ma2024segment} and Medical SAM 2~\cite{zhu2024medical}, are being leveraged as general‑purpose building blocks for diverse downstream imaging tasks.
This evolution illustrates a clear trajectory from CNN‑based architectures optimized for segmentation toward transformer‑enhanced hybrids that bridge local detail and global context, laying the groundwork for modern foundation models in medical imaging.




\subsection{Multimodal Foundation Models}
In recent years, foundational models in the medical domain have witnessed rapid development, with multimodal models achieving significant progress in cross-modal data fusion and representation learning. 
Since the release of CLIP~\cite{radford2021learning}, which demonstrated the power of large‑scale image–text contrastive pretraining, multimodal learning has emerged as a central research paradigm.
Building on this trend, Flamingo~\cite{alayrac2022flamingo} introduced a unified architecture for handling interleaved visual and textual inputs, advancing the design of scalable and versatile multimodal systems.

Translating these advances into the medical domain, MedCLIP~\cite{wang2022medclip} has emerged as a powerful tool for medical image–text alignment, substantially improving diagnostic assistance and electronic health record (EHR) analysis, and laying the groundwork for deep exploitation of multimodal medical data.
Multi-Granularity Cross-modal Alignment (MGCA) framework~\cite{wang2022multi} learns generalized medical visual representations by aligning medical images and radiology reports at pathological region‑level, instance‑levels, and disease‑level through cross‑attention and contrastive learning.
BioViL~\cite{boecking2022making} represents another key biomedical vision–language foundation model, improving image–report alignment through domain‑adapted semantic pretraining and enhanced cross‑modal representation learning.
To exploit the temporal structure of medical data, such as descriptions of disease progression commonly found in clinical reports, BioViL-T~\cite{bannur2023learning} extends this framework by incorporating temporal context from prior images and longitudinal reports, enabling more accurate and context‑aware performance across disease classification, phrase grounding, and report generation tasks.
Medical Spatio-Temporal (Med-ST)~\cite{yang2024unlocking} introduces a medical spatio‑temporal pretraining framework that jointly models the spatial relationships in multi‑view chest X‑rays (e.g., posteroanterior and lateral views) and the temporal progression captured in patients’ historical diagnostic records, thereby improving the capacity of vision–language models to leverage fine‑grained cross‑view and longitudinal information for enhanced multimodal representation learning.

\section{GenAI in Healthcare Applications}
\label{sec:applications}

The rapid evolution of foundation models has established GenAI as a cornerstone for advancing healthcare applications~\cite{shanmugam2025generative}.
Unlike traditional models optimized for narrow, task-specific objectives, GenAI leverages large-scale pretraining, cross-modal representation learning, and generative reasoning to address complex clinical needs. These capabilities make it particularly suitable for applications that require integrating heterogeneous biomedical data, generating context-aware outputs, and adapting to diverse clinical scenarios.
In practice, such strengths have driven progress across three pivotal domains: (i) Disease Diagnosis and Decision Support, where multimodal generative models unify imaging, physiological signals, and clinical narratives to support decision-making;
(ii) Medical Report Generation and Documentation, where language models automate the synthesis of structured and unstructured records;
and (iii) Drug Discovery and Biomedical Research, where molecular and protein language models accelerate hypothesis generation and compound design.
The following subsections review representative efforts in these areas, analyzing how generative and multimodal foundation models have been developed for modern healthcare applications.

\subsection{Disease Diagnosis and Decision Support}

Disease diagnosis presents a uniquely complex challenge in healthcare, requiring the integration of diverse information sources, ranging from imaging and physiological signals to clinical notes and longitudinal health records.
Recent advances in GenAI have driven a wave of diagnostic tools that move beyond traditional pattern recognition, enabling contextual reasoning and multimodal knowledge integration.

\noindent
\textbf{LLMs for Clinical Question Answering.}
Large Language Models (LLMs) have emerged as powerful tools for clinical question answering (QA), leveraging instruction tuning, retrieval augmentation, and chain‑of‑thought (CoT) reasoning to provide contextually grounded responses to complex medical queries.
By integrating vast biomedical corpora and aligning with domain‑specific instructions, these models are capable of understanding nuanced clinical questions, synthesizing knowledge across specialties, and generating evidence‑backed answers.

These methods can be broadly categorized into general‑purpose medical QA models~\cite{wang2023clinicalgpt,wang2024jmlr,wu2024pmc,singhal2023large,singhal2025toward,labrak2024biomistral} and specialized medical QA models~\cite{yu2023zero,li2024cancerllm,jia2024oncogpt,yang2024pediatricsgpt}.
General‑purpose models aim to provide broad coverage across diverse clinical topics, serving as versatile assistants for disease diagnosis and decision‑making.
BioMistral~\cite{labrak2024biomistral} leverages open‑source Mistral models~\cite{jiang2023mistral7b} with biomedical continual pretraining and instruction fine‑tuning to support broad‑coverage medical QA across general clinical topics.
JMLR~\cite{wang2024jmlr} employs a jointly trained retrieval‑augmented generation (RAG) framework, integrating LLMs with retrieval modules to access clinical guidelines and domain knowledge for medical QA.
These models are designed to operate as generalist assistants, capable of handling diverse medical queries across specialties to assist disease diagnosis and clinical decision-making.
Furthermore, specialized medical QA LLMs~\cite{yu2023zero,li2024cancerllm,jia2024oncogpt,yang2024pediatricsgpt} are being developed to address narrower clinical domains and patient populations.
For instance, PediatricsGPT~\cite{yang2024pediatricsgpt} is a Chinese pediatric assistant LLM fine‑tuned on pediatric guidelines and case‑based dialogues, providing tailored responses for pediatric care in Chinese clinical contexts.
These developments illustrate a continuum from broad‑coverage generalists to focused domain experts, highlighting how strategic pretraining and fine‑tuning align LLMs with the nuanced demands of clinical practice.

\noindent
\textbf{LVLMs for Diagnostic Reasoning.}
Through cross‑modal pretraining and alignment, VLMs are capable of integrating biomedical imaging with domain‑specific textual information, thereby providing interpretable, context‑aware diagnostic insights.
For example, LLaVA‑Med~\cite{li2023llava} offers a cost‑efficient framework for training vision‑language conversational assistants capable of answering open‑ended research questions about biomedical images, supporting flexible diagnostic exploration.

In addition, specialized VLMs have been developed for targeted diagnostic applications, with radiograph interpretation emerging as a major focus~\cite{thawkar2023xraygpt,liu2024g2d,zhang2023knowledge,lai2024carzero}.
Knowledge-enhanced Auto Diagnosis (KAD)~\cite{zhang2023knowledge} integrates structured medical domain knowledge into vision–language pretraining, aligning chest X‑ray images with knowledge‑grounded radiology report representations. By incorporating ontology‑based concept encoding and disease‑query mechanisms, KAD advances automated diagnosis for chest X-ray images by leveraging domain knowledge.
CARZero~\cite{lai2024carzero} replaces traditional cosine similarity-based alignment with cross‑attention mechanisms and incorporates an LLM‑driven prompt standardization strategy for zero‑shot radiology classification, achieving state‑of‑the‑art performance on chest radiograph diagnostic benchmarks.
Beyond radiology, VLMs are also being tailored to other clinical specialties.
For instance, OphGLM~\cite{gao2023ophglm} extends this paradigm to ophthalmology, aligning ocular imaging with domain‑specific knowledge for improved diagnostic reasoning in eye disease assessment,
while EchoCLIP~\cite{christensen2024vision} is an LVLM for echocardiography trained on over one million cardiac ultrasound videos and expert interpretations, enabling accurate cardiac function assessment and patient‑level reasoning across studies.

\noindent
\textbf{Agent‑Based Diagnostic Assistants.}
Agent‑based frameworks have recently emerged as a promising paradigm for disease diagnosis and clinical decision support, addressing the limitations of standalone LLMs by enabling multi‑agent collaboration~\cite{wang2025survey}.
In these methods, agents are assigned specialized roles, such as intent recognition, diagnostic reasoning, and treatment planning, to create a context‑aware and interactive healthcare delivery process.

A growing body of frameworks underscores the potential of multi‑agent systems in enhancing diagnostic reasoning and clinical decision‑making.
The multi‑agent system~\cite{dutta2024adaptive} simulates interactions between doctors and patients to refine diagnostic reasoning, demonstrating the effectiveness of multi‑agent collaboration in enhancing diagnostic performance.
Similarly, Ke et al.~\cite{ke2024mitigating} develop a multi‑agent framework that assigns diverse roles to LLM‑driven agents to mitigate cognitive biases in clinical decision‑making, significantly improving diagnostic accuracy in complex medical scenarios.
Medagent~\cite{tang2023medagents} proposes a multi‑disciplinary role‑playing framework in which LLM agents engage in multi‑round collaborative discussions, enhancing medical reasoning and decision‑making in zero‑shot diagnostic scenarios.
MDagents~\cite{kim2024mdagents} introduces an adaptive collaboration framework that dynamically assigns solo or group roles to LLM agents based on task complexity.
Moreover, Agent Hospital~\cite{li2024agent} simulates the full clinical workflow using evolvable LLM-powered agents, with its MedAgent-Zero paradigm enabling doctor agents to iteratively improve through simulated and real-world experience, demonstrating strong generalization to real-world diagnostic tasks.

Beyond general diagnostic assistants, several frameworks target specialized clinical domains, demonstrating the adaptability of multi‑agent systems to specific healthcare tasks.
ClinicalAgent~\cite{yue2024clinicalagent} integrates large‑scale domain knowledge with multi‑agent collaboration to enhance clinical trial outcome prediction.
Polaris~\cite{mukherjee2024polaris} focuses on real‑time patient–AI clinical dialogues, using a safety‑oriented constellation of primary and specialist agents trained on clinical care plans, regulatory documents, and simulated dialogues.
These agent‑based systems represent a shift from monolithic LLMs to dynamic, role‑oriented multi‑agent systems, enhancing reasoning and embedding domain expertise within clinical workflows.

\subsection{Medical Report Generation and Documentation}

Medical report is a cornerstone of clinical care but remains a cognitively and time‑consuming demanding task for healthcare providers.
GenAI offers the potential to streamline documentation by automating the synthesis of structured and unstructured information into coherent, contextually relevant reports, thereby reducing administrative burden and enhancing communication among clinicians and with patients.
Unlike models designed for disease diagnosis, which focus on classification or decision support, report generation models prioritize producing detailed content with specialized medical terminology that documents clinical findings~\cite{liu2025enhanced}.

\noindent
\textbf{Vision–Language Encoder–Decoder Pipelines.}
Early approaches to report generation primarily leverage encoder–decoder architectures that combine a visual feature extractor with a text generation module.
In this paradigm, a vision encoder (e.g., ResNet~\cite{huang2023kiut,yang2023radiology}, ViT~\cite{nicolson2023improving,wang2023metransformer}) is used to transform medical images such as X-rays, CT, and pathology into high-dimensional feature representations.
These representations are then passed to a language decoder that generates descriptive reports in natural language.
Several notable studies exemplify this paradigm.
R2Gen~\cite{chen2020generating} adopts a memory-driven Transformer that allows the decoder to revisit visual features, improving consistency and coverage in generated reports.
Zhang et al.~\cite{zhang2020radiology} extend the encoder–decoder pipeline by incorporating a graph convolutional module over disease findings, enabling explicit modeling of relationships between abnormalities (e.g., effusion and cardiomegaly) and yielding more clinically accurate reports.
These methods established the foundational blueprint for report generation, while their limitations in factual accuracy and clinical coverage motivated the shift toward more advanced multimodal and agent-based approaches.

\noindent
\textbf{End-to-End Multimodal Large Language Models.}
Unlike earlier encoder–decoder pipelines, recent efforts pursue end-to-end multimodal large language models (MLLMs) that unify visual and textual representations within a single framework.
These models aim to achieve tighter integration between medical image understanding and language generation, thereby reducing reliance on handcrafted fusion strategies and enabling more flexible downstream applications.
For instance, XrayGPT~\cite{thawkar2023xraygpt} bridges the gap between general-purpose VLMs and radiology by aligning a medical visual encoder (MedClip) with a fine-tuned LLM (Vicuna) and introducing 217k interactive summaries from free-text radiology reports, thereby equipping the model with strong visual–textual grounding.
MRM~\cite{zhou2023advancing}  unifies radiograph understanding and report comprehension through masked record modeling, jointly reconstructing image patches and report tokens to learn transferable, knowledge-enhanced vision–language representations.
To further improve clinical accuracy in report generation, researchers have incorporated diverse strategies such as adaptive patch–word alignment for improved explainability~\cite{chen2024fine}, training paradigms that promote deeper incorporation of visual features into LLMs~\cite{lee2024llm}, and multi-view longitudinal learning with patient-specific priors~\cite{liu2025enhanced}.
These innovations underscore the trend toward end-to-end multimodal reasoning, advancing both visual–textual alignment and the clinical reliability of generated reports.


\noindent
\textbf{Agent-based Medical Report Generation.}
More recently, agent-based methods represent a rapidly emerging paradigm in the medical domain, emphasizing dynamic reasoning, planning, and interactive workflows with external tools and knowledge bases.
In the context of radiology and broader medical reporting, agent-based systems can function as autonomous assistants that query structured resources (e.g., UMLS, RadLex, or EHR data), validate generated findings against medical guidelines, and even coordinate multi-step reasoning pipelines.
Multimodal Multi-Agent RRG~\cite{yi2025multimodal} adopts a modular design that mirrors the clinical reasoning workflow, deploying specialized agents for retrieval, drafting, visual analysis, refinement, and synthesis.
RadCouncil~\cite{zeng2024enhancing} follows a council-style architecture, where agents sequentially retrieve similar cases, generate impressions, and review outputs, ensuring that final reports achieve higher clarity and diagnostic reliability.
To enhance explainability, CBM-RAG~\cite{alam2025cbm} integrates interpretable concept bottleneck models with a multi-agent retrieval-augmented generation process, enabling reports that are grounded in explicit clinical evidence. MRGAgents~\cite{wang2025mrgagents} introduces disease-specific agents trained on category-focused subsets of data, effectively mitigating the tendency of models to over-report normal findings and improving the comprehensiveness of abnormality detection.
Although still at an early stage, agent-based methods highlight a promising direction, integrating the reasoning flexibility of autonomous agents with the domain expertise of medical knowledge bases to achieve more reliable, transparent, and clinically useful report generation.

\subsection{Drug Discovery and Biomedical Research}

Drug discovery is a complex, multi-stage pipeline that spans from target identification and validation to lead compound generation, optimization, and clinical translation.
Traditional approaches often require more than a decade of effort and substantial financial investment, with high attrition rates during clinical trials. Recent GenAI approaches are reshaping this paradigm, ranging from task-specific methods for individual pipeline stages to cross-stage agent-based frameworks that enable end-to-end workflows.

\noindent
\textbf{Task-Specific Drug Discovery Methods.}
A growing body of work leverages LLMs to enhance performance in specific drug discovery subtasks.
For example, LMFL~\cite{brahmavar2024generating} with logical feedback refines molecular generation through constraint-guided reasoning, improving the quality of novel pharmacological leads.
DrugAgent~\cite{liu2024drugagent} formulates drug–target interaction prediction as an agent-driven task, where an LLM planner collaborates with domain-specific tools to iteratively search for high-performing models.
DrugAssist~\cite{ye2025drugassist} introduces an interactive optimization model that employs human–AI dialogue to refine molecular properties across multiple objectives, thereby advancing molecule optimization.
CLADD~\cite{lee2025rag} further extends this paradigm by adopting a multi-agent RAG framework that integrates biomedical knowledge graphs for context-aware molecule evaluation.

\noindent
\textbf{Cross-Stage and End-to-End Agent Frameworks.}
Beyond single-task improvements, recent efforts emphasize building generalizable orchestration systems that span multiple pipeline stages. DrugPilot~\cite{li2025drugpilot} represents such an approach by introducing a parameterized memory pool to manage heterogeneous biomedical data, enabling multi-stage reasoning and efficient tool coordination for generation, optimization, and property prediction. Similarly, Solovev et al.~\cite{solovev2024towards} decompose complex discovery tasks into manageable subtasks, with planner, validator, and tool-calling agents coordinating pretrained generative models, docking algorithms, and predictive modules.
These systems achieve near end-to-end automation, demonstrating superior performance in complex real-world drug discovery scenarios, such as neurodegenerative disease research.



\subsection{Multifunctional Generalist Models}


Beyond task-specific architectures, a new class of generalist foundation models has emerged, aiming to unify diverse biomedical modalities and tasks within a single framework.
These multifunctional systems are designed to handle a wide spectrum of applications, including open-QA, closed-QA, and report or image generation, by jointly leveraging domain-specific medical knowledge and multimodal reasoning capabilities.
BiomedGPT~\cite{luo2023biomedgpt} represents a transformative approach that connects biological processes with human communication, facilitating enhanced comprehension of essential biological mechanisms within medical contexts.
HuatuoGPT~\cite{chen2024huatuogpt} introduces PubMedVision, a collection of 1.3 million high-quality medical specimens that substantially enhances flexibility and performance across medical vision–language tasks.
Similarly, HealthGPT~\cite{lin2025healthgpt} presents a unified framework for medical vision-language comprehension and synthesis, employing a heterogeneous knowledge adaptation methodology that yields consistent performance gains across diverse applications.

Expanding to specialized diagnostic domains, EyecareGPT~\cite{li2025eyecaregpt} incorporates an adaptive resolution system for dynamic optimization alongside a layer-wise dense connector that effectively integrates multi-scale visual characteristics. These innovations reveal considerable promise for advancing open research initiatives in automated ophthalmological diagnostics.
HeartcareGPT~\cite{xie2025heartcare} develops an extensive multimodal foundation architecture featuring an innovative bidirectional ECG abstract tokenization methodology, enabling detailed electrocardiogram analysis and interpretation capabilities.
Overall, these models illustrate the evolution of intelligent healthcare from single-modality to multimodal paradigms, from single-model to multi-agent frameworks, and from single-task functionality to truly multifunctional generalist systems.



\section{On the Design of a Data-Centric Ecosystem}
\label{sec:discussion}

In this section, we discuss the current challenges and delineate the key features required for a data-centric ecosystem to support GenAI for healthcare.

\subsection{Challenges}
Recent years have witnessed an unprecedented growth in both the volume and heterogeneity of healthcare data, spanning structured EHRs, unstructured medical imaging, physiological waveforms, and emerging omics profiles. Coupled with advances in GenAI, these rich multimodal resources hold transformative potential for clinical decision support, medical documentation, patient engagement, and biomedical discovery.
While these approaches demonstrate significant potential, most remain confined to research prototypes and struggle to achieve real-world deployment.
For example, a model that generates echocardiography summaries from imaging and waveform data may perform well on a local research dataset, but fail in deployment due to incompatible data formats, lack of reliable provenance tracking, or the absence of feedback loops to adapt to evolving clinical protocols.
Through a systematic analysis of existing efforts across diverse healthcare applications, we identify three interlocking challenges rooted in the data ecosystem, spanning static data fragmentation, inadequate dynamic management, and the lack of infrastructures for data–model co-evolution.
These structural barriers form a critical bottleneck in translating GenAI from proof-of-concept to safe, accurate, and broadly accessible clinical practice.

\subsubsection{Data Fragmentation and Heterogeneity}
Healthcare data is inherently fragmented and heterogeneous, spanning structured EHR tables with longitudinal lab values and coded diagnoses; unstructured clinical narratives in discharge summaries; high-dimensional imaging modalities such as MRI, CT, and histopathology slides; continuous physiological waveforms from ICU monitors; and emerging omics datasets from genomics and proteomics.
Beyond raw data, derived resources, such as semantic embeddings for retrieval-augmented generation (RAG) and domain-specific medical knowledge graphs, further diversify the ecosystem.
These assets are often stored in incompatible formats across institutions and governed by inconsistent metadata schemas, with varying degrees of data completeness, quality, and annotation fidelity depending on institutional standards and device calibration.
Multi-institutional datasets therefore exhibit uneven data quality and contribution balance, as some organizations supply larger or cleaner subsets while others contribute sparse or noisy samples.
Such fragmentation appears not only as multi-format inconsistency but also as semantic, structural, and temporal discontinuities that disrupt knowledge alignment and longitudinal reasoning across modalities.

For GenAI systems, these discontinuities propagate along the entire pipeline.
At the \textit{input level}, fragmented data reduces contextual completeness. For instance, a RAG module may retrieve laboratory results without corresponding imaging evidence, yielding incomplete prompts.
At the \textit{representation level}, modality-specific preprocessing and inconsistent annotations distort the embedding space, hindering cross-modal alignment and leading to biased generation.
At the \textit{generation level}, incoherent metadata or missing temporal links cause factual inconsistencies in generated clinical summaries, such as reporting a treatment before its documented prescription.
Ultimately, data fragmentation undermines GenAI's ability to form a unified patient-centric understanding, erodes factual faithfulness, and limits model transferability across clinical sites.

\subsubsection{Data Lifecycle Management and Governance}
Effective GenAI in healthcare relies not only on data availability but also on the continuous management of data across its entire lifecycle, from acquisition and curation to utilization, monitoring, and archival.
In practice, most institutions adopt static, one-off pipelines where data are ingested, cleaned, and frozen for model training, without systematic procedures for versioning, quality auditing, or periodic renewal. Updates to clinical protocols, device calibration standards, or labeling criteria are seldom propagated to existing datasets, resulting in temporal inconsistencies and concept drift across data vintages.
Recent consensus efforts in diagnostic AI~\cite{sounderajah2025stard} underscore the importance of documenting data acquisition protocols, preprocessing steps, and dataset partitioning to ensure transparency and reproducibility. These principles illustrate how inadequate lifecycle management undermines traceability and regulatory assurance.

Deficiencies in data governance further amplify these lifecycle challenges. Inconsistent anonymization practices and heterogeneous access-control policies hinder secure data sharing across institutions, while the absence of unified provenance tracking impedes accountability and interpretability.
Prior analyses of clinical AI governance frameworks have highlighted similar gaps in transparency and traceability~\cite{moor2023foundation,shanmugam2025generative}.
Without continuous stewardship and harmonized oversight, generative systems rely on outdated, weakly contextualized information, compromising both factual reliability and compliance with evolving ethical and regulatory standards.


\subsubsection{Data–Model Co-Evolution Infrastructure}
Current GenAI pipelines in healthcare remain largely unidirectional: data flow into model training, yet little information flows back to refine the underlying datasets.
Consequently, models drift away from the realities they were designed to represent, while data pipelines remain passive recipients rather than active participants in continuous improvement.
Empirical assessments of large hospital data networks, such as MIMIC-III~\cite{johnson2016mimic} and the eICU Collaborative Research Database~\cite{pollard2018eicu}, reveal that record revisions, late entries, and retrospective corrections are propagated only partially or after prolonged delays (often exceeding several months), illustrating how the absence of responsive infrastructure prevents timely synchronization between data and model updates~\cite{lewis2023electronic}.
Ideally, this feedback loop would enable models to surface systematic errors or underrepresented patterns, prompting targeted data enrichment or fine-grained re-training. For instance, a radiology report generation model that identifies misinterpretations of rare fracture types could initiate both dataset augmentation for such cases and architectural refinement for improved spatial reasoning.
Without such adaptive co-evolution, GenAI performance decays over time, contextual relevance erodes, and clinical adoption stalls.


\subsection{Key Features}

Beyond addressing present challenges, a data-centric ecosystem for generative healthcare AI should be envisioned as a proactive, evolving foundation that treats data as a dynamic substrate for adaptive, trustworthy, and clinically aligned intelligence.
We distill this perspective into four interdependent features:

\begin{itemize} [leftmargin=0.4cm]
\item \textbf{Organized Multimodal Representation and Interoperability.}
At the core of the ecosystem lies the ability to reconcile the profound heterogeneity of healthcare data.
Clinical knowledge is distributed across diverse modalities, from structured EHR tables and radiological images to continuous physiological signals and emerging omics profiles, each encoded in fundamentally different formats and semantics.
To transform these fragmented resources into a coherent whole, data are systematically organized under standardized schemas, then embedded into shared semantic representation spaces.
Organized in this manner, multimodal data can be consistently indexed, queried, and linked, establishing a coherent substrate on which higher-level reasoning and model adaptation can be built.

\item \textbf{Semantic Enrichment and Contextual Retrieval.}
Building on organized, interoperable data, the ecosystem derives semantically enriched assets that expose latent clinical relationships and temporal dependencies.
Raw records are transformed into higher-order abstractions such as embedding vectors and temporal knowledge graphs, which capture latent associations across patients, diseases, and modalities.
These enriched resources underpin contextual retrieval, enabling generative models to access task-relevant evidence dynamically during inference.
By grounding outputs in semantically curated and clinically aligned information, enrichment and retrieval mechanisms support RAG, cross-modal diagnostic reasoning, and personalized decision support.

\item \textbf{Adaptive Feedback and Co-Evolution.}
A defining feature of a sustainable ecosystem is the capacity to learn continuously from its own use. Feedback signals arising from clinician interactions, model performance monitoring, and drift detection are systematically reintegrated into both data pipelines and model refinement through concrete mechanisms such as re-indexing, relabel audits, and PEFT refresh.
This feedback loop transforms the relationship between data and models from a one-way dependency into a bidirectional process of co-evolution: enriched datasets enhancing model generalization while model-driven insights guide targeted data curation.

\item \textbf{Trustworthy and Federated Governance.}
Trust is embedded as an infrastructural property of the ecosystem through privacy-preserving computation, provenance tracking, and continuous regulatory compliance.
Transparent auditability ensures confidence in data use and model outputs, while federated infrastructures, such as our own FALCON \cite{falcon}, enable multi-institutional collaboration without centralizing sensitive records.
In combining trustworthiness with federated design, the ecosystem balances large-scale knowledge sharing with local custodianship, supporting sustainable adoption in clinical practice.

\end{itemize}

\section{Perspectives}
\label{sec:perspectives}

\subsection{Design of \oursys}
\begin{figure}[t]
    \centering 
\includegraphics[width=1\linewidth]{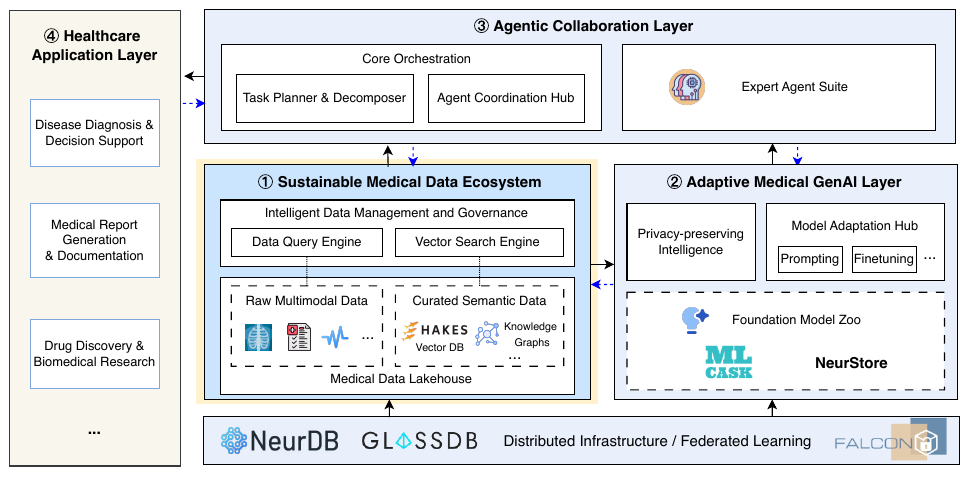} 
    \caption{\oursys Architecture.
    }
    \label{fig:framework}
\end{figure}

To bridge the persistent gap between GenAI prototypes and real-world clinical workflows, we propose \oursys, a \textbf{\underline{S}}ustainable, \textbf{\underline{A}}daptive, and \textbf{\underline{G}}enerative \textbf{\underline{E}}cosystem for Healthcare.
The conceptual framework of \oursys is illustrated in Figure~\ref{fig:framework}.
\oursys reconceptualizes the role of data from a static, peripheral resource to a continuously evolving, intelligent substrate that co-develops in synergy with generative models.
The framework integrates three core components: (1) Sustainable Medical Data Ecosystem, (2) Adaptive Medical GenAI Layer, and (3) Agentic Collaboration Layer. These three components operate in concert to underpin the (4) Healthcare Application Layer, enabling scalable downstream tasks such as disease diagnosis, medical report generation, and drug discovery.
By doing so, \oursys establishes a data-centric pathway for translating GenAI from controlled demonstrations to safe, effective, and widely adopted clinical practice.

\subsection{Sustainable Medical Data Ecosystem}
The sustainable medical data ecosystem forms the foundational layer of \oursys, designed to restructure fragmented medical information into a unified and governable resource.
It functions as an evolving infrastructure that curates, manages, and governs healthcare data to sustain GenAI's reliability and clinical relevance over time.
\blue{We categorize heterogeneous medical data into two interdependent tiers, raw multimodal data (e.g., EHRs, imaging, and physiological signals) and curated semantic data (e.g., embedding vectors and medical knowledge graphs) derived from it.}
To accommodate both, we propose a two-tier \textbf{Medical Data Lakehouse} architecture that preserves the full fidelity of original clinical records while enriching them into machine-interpretable formats.
Built on top of this foundation is \textbf{Intelligent Data Management and Governance}, whose core consists of \blue{a data query engine for structured and unstructured access to the aforementioned raw data tier, and a vector search engine for high-dimensional retrieval across multimodal embeddings within the semantic data tier.}
These engines operate in concert to unify multimodal ingestion, semantic enrichment, and cross-modal retrieval.
In healthcare contexts, governance is critical to ensuring continuous privacy compliance with regulations such as HIPAA\footnote{Health Insurance Portability and Accountability Act, Available at: \url{https://www.hhs.gov/hipaa}} and GDPR\footnote{General Data Protection Regulation, Available at: \url{https://gdpr-info.eu}}, along with provenance tracking that links every model output to its originating data, processing steps, and consent records, maintaining a verifiable chain of custody from data acquisition to model output.
Such measures not only safeguard patient confidentiality but also build trust in GenAI outputs, enabling their safe and sustained integration into clinical workflows.

The Sustainable Medical Data Ecosystem functions as the living backbone of \oursys, continuously ingesting, harmonizing, and semantically enriching diverse healthcare data streams. Its operation is interwoven with the other architectural layers, enabling both data provision and feedback-driven refinement.
With the Adaptive Medical GenAI Layer, the ecosystem delivers high-quality, context-aware inputs for model training, prompting, and fine-tuning. In return, it receives model-driven feedback (e.g., error patterns, uncertainty signals, drift detection) to guide targeted data enrichment and pipeline adjustments.
In the Agentic Collaboration Layer, the ecosystem underpins retrieval-augmented generation (RAG) workflows by providing agents with integrated multimodal retrieval capabilities. Agents can query across structured, unstructured, and high-dimensional semantic representations, ensuring that decision-making is consistently grounded in comprehensive, up-to-date evidence.
The Healthcare Application Layer closes the feedback loop by indirectly contributing enriched real-world signals back into the data ecosystem. Clinical applications not only consume model outputs but also produce high-value feedback in the form of user interactions, clinician corrections, and task outcome records. These signals flow through the Agentic Collaboration Layer, where they are contextualized with task metadata and provenance details before re-entering the data ecosystem. Here, the Intelligent Data Management and Governance component performs quality control (e.g., cleaning, labeling), compliance verification, semantic integration, and re-insertion into the medical data lakehouse.
Through these coordinated interactions, the Sustainable Medical Data Ecosystem operates not merely as a data provider but as a strategic orchestrator, governing the flow, integrity, and contextual richness of information across \oursys. This sustainable design transforms data from a static asset into a continuously improving, governance-compliant foundation for adaptive and clinically aligned GenAI.

\subsection{Adaptive Medical GenAI Layer}
The Adaptive Medical GenAI Layer forms the intelligence core of \oursys, enabling generative models to continuously learn, adapt, and operate in alignment with evolving clinical contexts.
Its primary role is to leverage the high-quality, context-aware data curated by the Sustainable Medical Data Ecosystem to build and adapt models for a wide spectrum of specialized healthcare tasks, ensuring that generative intelligence remains both clinically relevant and task-specific.

At its foundation lies the \textbf{Foundation Model Zoo}, a curated collection of large-scale pre-trained models spanning multiple modalities, such as LLMs for text and MLLMs for imaging, signals. These models serve as the backbone for downstream adaptation, enabling rapid deployment across varied clinical tasks without training from scratch.
Built atop this foundation is the \textbf{Model Adaptation Hub}, which provides a flexible toolkit for tailoring models to specific healthcare scenarios.  
\blue{This includes zero- and few-shot prompt engineering for rapid task generalization, parameter-efficient fine-tuning (PEFT) such as LoRA~\cite{hu2022lora} and extra adapters for domain-specific optimization, and full supervised fine-tuning (SFT) and architecture-level modifications when deeper specialization is required.}
The hub also integrates real-world feedback from the application layer, allowing continuous refinement through incremental learning and drift-aware re-training strategies.
Complementing these adaptation capabilities is the \blue{\textbf{Privacy-Preserving Intelligence} subsystem, which embeds privacy-by-design mechanisms into all stages of model lifecycle management. Techniques such as differential privacy, zero-knowledge proof (ZKP)~\cite{zkpooi24}, and secure multi-party computation (SMPC) ensure that sensitive patient data remains decentralized while still contributing to collective model improvement, enabling secure model adaptation and inference in compliance with stringent healthcare regulations.
}

Operationally, the Adaptive Medical GenAI Layer interacts closely with the Agentic Collaboration Layer by exposing adaptable model endpoints that agents can query, orchestrate, and chain within complex inference workflows.
Through this interface, agents dynamically select optimal models and inference strategies, enabling the delivery of accurate, context-aware outputs to a wide range of downstream tasks in the Healthcare Application Layer.
Crucially, application-level user interactions, clinical corrections, and performance metrics are routed back through the agentic layer, enriched with provenance and task metadata, and used to trigger incremental updates or re-training for model refinement.
In this way, the Adaptive Medical GenAI Layer not only serves as the execution engine for healthcare intelligence but also as a self-optimizing module, aligning model behavior with evolving clinical practices and application requirements.

\begin{figure}[t]
    \centering 
\includegraphics[width=1\linewidth]{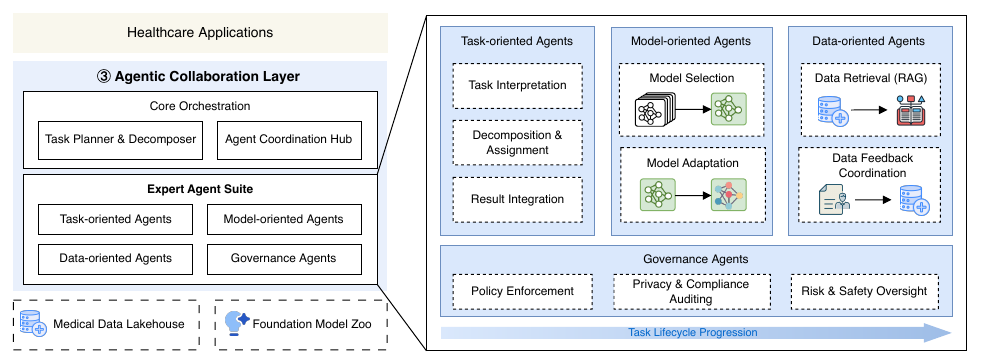} 
    \caption{Overview of the Agentic Collaboration Layer.}
    \label{fig:agentic_layer}
\end{figure}

\subsection{Agentic Collaboration Layer}
The Agentic Collaboration Layer serves as the cognitive coordination hub of the \oursys architecture, bridging adaptive generative intelligence with real-world healthcare applications. It transforms raw model capabilities into task-oriented, context-aware actions through dynamic orchestration of specialized agents, as illustrated in Figure~\ref{fig:agentic_layer}.
By decomposing complex clinical objectives into executable subtasks, assigning them to the appropriate agents, and integrating results into coherent solutions, this layer enables flexible and auditable decision-making.

Specifically, the \textbf{Core Orchestration} component governs this layer, with a Task Planner and Decomposer that interprets high-level clinical objectives and decomposes them into well-defined subtasks with explicit dependencies, and an Agent Coordination Hub that dynamically allocates these subtasks to optimal agents in the \textbf{Expert Agent Suite}, ensuring seamless cooperation among diverse agent types.
In particular, the Expert Agent Suite comprises four specialized agent classes, each designed to address distinct operational dimensions.
First, the Task-oriented Agents interpret domain-specific objectives, decompose workflows into executable sequences, and integrate partial results into coherent final outputs. They act as the bridge between abstract clinical intent and concrete multi-step execution plans.
Second, the Model-oriented Agents manage the end-to-end lifecycle of GenAI model utilization. They select the most suitable model from the Foundation Model Zoo and perform model adaptation using techniques from the Model Adaptation Hub to meet evolving healthcare needs.
Third, the Data-oriented Agents handle the operations with the Sustainable Medical Data Ecosystem, performing RAG from the Medical Data Lakehouse to provide context-rich inputs for downstream tasks and coordinating feedback integration to enhance data quality and semantic consistency.
Additionally, the Governance Agents safeguard the safety boundaries of the system throughout the entire task lifecycle progression. These agents sequentially enforce policies, audit privacy and regulatory compliance, and oversee risks, ensuring system safety and adherence to clinical, legal, and institutional standards across the task lifecycle.

In operation, tasks from the Healthcare Application Layer are decomposed, allocated, and executed across agents, with outputs validated and integrated into coherent clinical responses.
Two complementary feedback loops flow through the Agentic Collaboration Layer: (i) data feedback, in which user interactions, clinical corrections, and contextual task information are routed back to enrich and update the Sustainable Medical Data Ecosystem, and (ii) model feedback, where performance metrics, error patterns, and drift signals guide targeted model adaptation in the Adaptive Medical GenAI Layer.
Together, these loops enable continuous refinement of both data pipelines and model behaviors, ensuring \oursys remains responsive to the evolving healthcare demands.

\begin{figure}[t]
    \centering 
\includegraphics[width=1\linewidth]{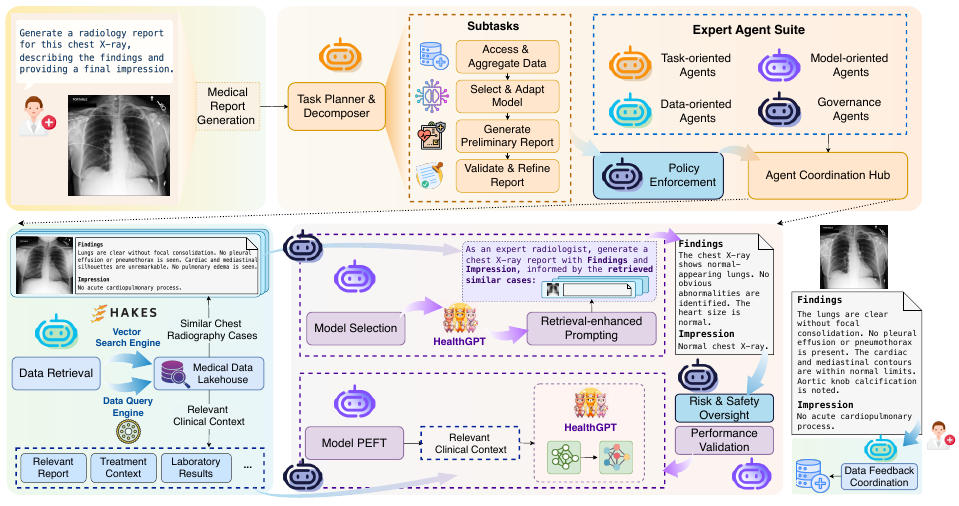
} 
    \caption{Workflow of \oursys for Radiology Report Generation from Chest X-rays. }
    \label{fig:case_study}
\end{figure}

\subsection{\oursys in Action}
\subsubsection{Workflow Illustration}
To illustrate the operational dynamics of \oursys, we consider a representative clinical scenario: generating a medical report from a chest X-ray image.
Chest X-ray report generation is a core diagnostic task in radiology, requiring the transformation of complex medical images into structured, clinically actionable reports that integrate visual findings with patient context.
The end-to-end workflow of this process is illustrated in Figure~\ref{fig:case_study}, which demonstrates how \oursys coordinates data retrieval, model selection, and expert agents to produce accurate and contextually grounded radiology reports, with the input chest X-ray image randomly selected from the MIMIC-CXR dataset~\cite{johnson2019mimic}.

In \oursys, this task begins in the \textbf{Healthcare Application Layer}, where a clinician submits a chest X-ray
through the Medical Report Generation \& Documentation interface.
The request is then relayed to the \textbf{Agentic Collaboration Layer}, where the Task Planner \& Decomposer leverage the task-oriented agents in the Expert Agent Suite to interpret the clinical objective and formulate a multi-stage plan:
(1) accessing and aggregating the patient-specific context,
(2) selecting and adapting an optimal vision-language model for chest radiography,
(3) generating a preliminary report,
and (4) validating and finalizing the report.
The Agent Coordination Hub subsequently orchestrates the execution of these subtasks by delegating them to the appropriate Expert Agents within the suite, ensuring that model-oriented, data-oriented, and governance functions are seamlessly integrated throughout the process.

Specifically, the Data-oriented Agents access the \textbf{Sustainable Medical Data Ecosystem}, leveraging Vector Search Engine HAKES~\cite{hakes} to retrieve semantically similar chest radiography cases from curated archives.
The retrieved data are then routed to the Model-oriented Agents, which invoke the \textbf{Adaptive Medical GenAI Layer} to select an appropriate vision–language foundation model from the Foundation Model Zoo, such as HealthGPT~\cite{lin2025healthgpt}.
After an appropriate model is selected, targeted adaptation strategies in the model adaptation hub are realized through RAG, a retrieval-enhanced prompting method that incorporates retrieved chest X-ray exemplars into the inference stage to guide draft report generation.
Following this step, the Model-oriented Agents conduct performance validation, which may trigger targeted updates of the model. To enable such updates, the Data-oriented Agents query the \textbf{Sustainable Medical Data Ecosystem} via the Data Query Engine to extract relevant clinical context (e.g., longitudinal imaging records, prior radiology reports, and pertinent EHR segments). Leveraging this context, targeted adaptation strategies such as PEFT are employed to optimize model performance for the present case.
The updated model is then reintegrated into the inference workflow to generate a refined medical report.

The Privacy-Preserving Intelligence subsystem safeguards patient confidentiality during both model adaptation and inference in chest X-ray report generation.
For fine-tuning, raw chest X-rays and clinical histories remain within their originating institutions, where Model-oriented Agents conduct local feature extraction and parameter updates inside secure execution environments. Only anonymized representations and encrypted weight deltas are transmitted for secure aggregation across sites, enabling model refinement without exposing identifiable imaging or EHR data.
During the inference process, the patient's X-ray is transformed into privacy-preserving embeddings on-site, ensuring secure model improvement and report generation.
Throughout this process, the Governance Agents operate continuously across the task lifecycle: enforcing policy compliance during task planning, auditing privacy and regulatory adherence during data retrieval and model adaptation, and performing risk and safety oversight before the report is released.

This workflow is reinforced by two complementary feedback loops.
The model feedback loop is activated through task-level performance indicators, such as diagnostic accuracy for rare thoracic findings, and enables targeted model updates in the Adaptive Medical GenAI Layer.
In parallel, a data feedback loop operates downstream, where clinicians’ edits and diagnostic confirmations flow from the Healthcare Application Layer through the Agentic Collaboration Layer’s Data-oriented Agents, which enrich them with task-specific context, and reintegrate the curated information into the Sustainable Medical Data Ecosystem to enhance future retrieval and semantic curation.
Through this multi-layer collaboration, \oursys delivers high-quality, contextually grounded chest X-ray reports that are accurate, compliant, and continuously improving.

\begin{figure}[t]
    \centering 
\includegraphics[width=0.8\linewidth]{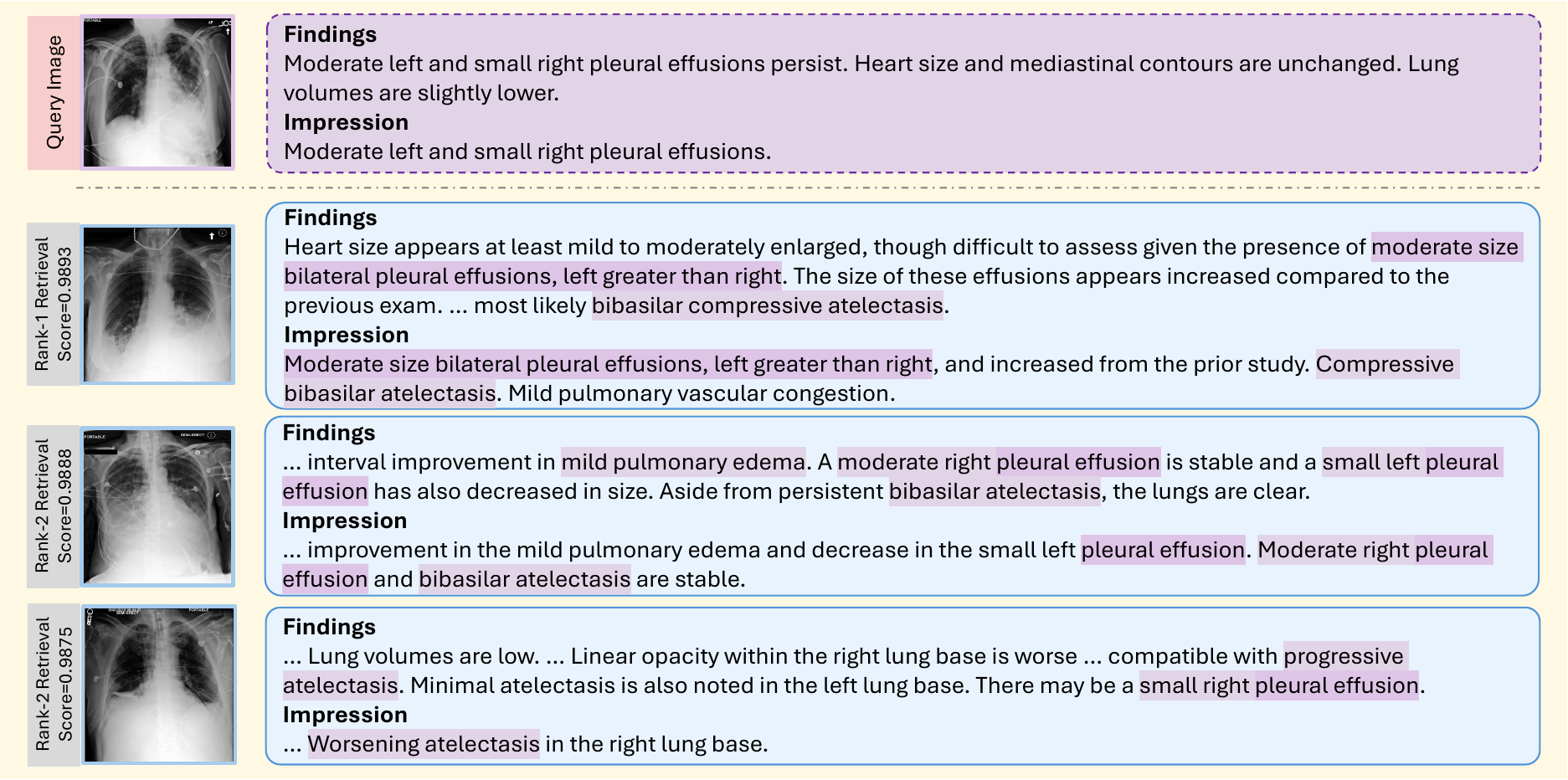}
    \caption{Retrieval case study illustrating query image and top retrieved samples, implemented with HAKES~\cite{hakes} using BioViL-T~\cite{bannur2023learning} as the vision encoder. }
\label{fig:case_study_retrieval}
\end{figure}

\begin{figure}[t]
    \centering 
\includegraphics[width=1\linewidth]{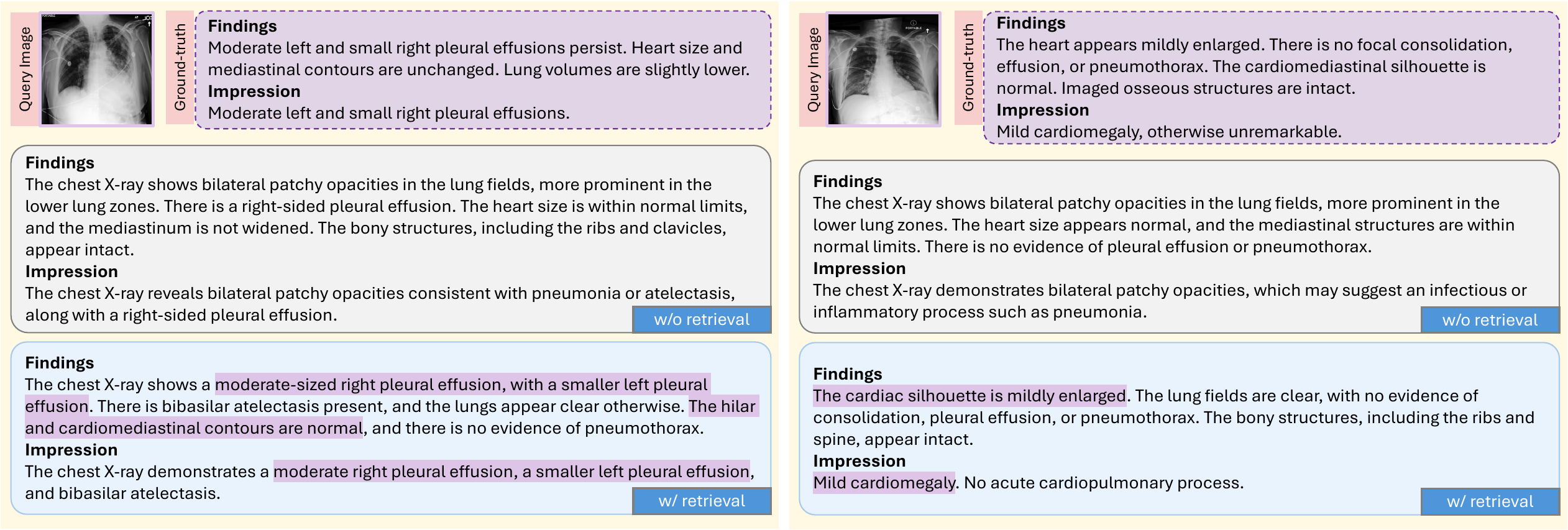}
    \caption{Case study of chest X-ray report generation with and without retrieval-augmented generation, using HealthGPT~\cite{lin2025healthgpt} as report generation backbone and incorporating top-3 retrieved cases.}
\label{fig:case_study_retrieval_compare}
\end{figure}

\subsubsection{Case Studies on Functional Capabilities of \oursys}
To further validate the practical value of \oursys, we present case studies in chest X-ray report generation. Through functional demonstrations of retrieval-augmented generation and data–model co-evolution via adaptive feedback, we illustrate how \oursys transforms data into actionable evidence and supports continuous refinement of generative outputs.

In radiology, the ability to ground generative reports in prior clinical evidence is essential for both accuracy and trust.
Within \oursys, retrieval fulfills this role by surfacing the most relevant chest X-ray–report pairs from the MIMIC-CXR archive, returning a panel of top-ranked images and reports that highlight overlapping findings such as “pleural effusion” (Figure~\ref{fig:case_study_retrieval}) and provide concrete precedents for downstream generation.
When integrated into generation, this evidence directly reshapes the output. As illustrated in Figure~\ref{fig:case_study_retrieval_compare}, baseline reports without retrieval tend to rely on generic or partially incomplete descriptions, whereas retrieval-augmented outputs incorporate precise, clinically meaningful phrases that are directly traceable to retrieved examples.

\begin{figure}[t]
    \centering 
\includegraphics[width=1\linewidth]{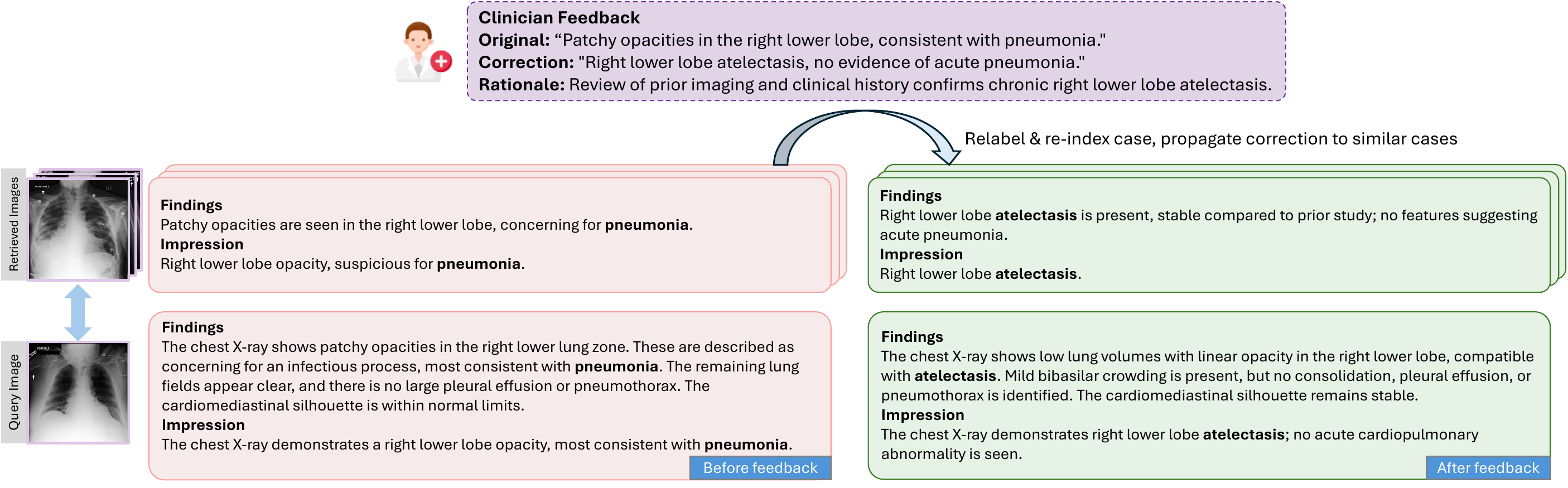}
    \caption{Case study on data–model co-evolution via adaptive feedback. }
\label{fig:case_study_feedback}
\end{figure}

\oursys further advances functionality by operationalizing clinician feedback as a driver of continuous data–model co-evolution.
A common clinical scenario arises when an opacity initially read as suspicious for pneumonia is later corrected to atelectasis based on prior imaging and clinical history.
In \oursys, such feedback is not treated as an isolated correction but is systematically integrated into the ecosystem: the case is relabelled, re-indexed, and linked to similar cases for further verification and correction.
By reshaping the retrieval context, the system ensures that when the same image is reprocessed, the report moves from an uncertain pneumonia diagnosis to a precise description of atelectasis, as illustrated in Figure~\ref{fig:case_study_feedback}.
This demonstration highlights how adaptive feedback closes the loop between clinicians, data infrastructure, and generative models, enabling practical data–model co-evolution that enhances factual accuracy and clinical trust without requiring full retraining.

\section{Conclusions}
\label{sec:conclusion}

Generative Artificial Intelligence (GenAI) holds immense promise for healthcare, but its clinical impact will remain limited without rethinking the data foundations that sustain it.
In this paper, we propose a data-centric paradigm that positions medical data ecosystems as the foundational substrate for generative healthcare systems.
We present \oursys, a sustainable, adaptive, and generative framework that tackles three interlocking challenges: data fragmentation, data life cycle governance, and data–model co-evolution, enabling GenAI to progress from proof-of-concept to reliable, scalable deployment. By exploiting data-centric GenAI systems as the engine of progress, we envision a future in which GenAI becomes a trustworthy tool in delivering truly transformative healthcare.

{\footnotesize
\bibliographystyle{unsrtnat}
\bibliography{reference}
}

\end{document}